\documentclass[a4paper, 10pt, conference]{ieeeconf}      
\usepackage{FG2021}

\FGfinalcopy 

\IEEEoverridecommandlockouts                             
\usepackage{graphics} 
\usepackage{epsfig} 
\usepackage{times} 
\usepackage{amsmath} 
\usepackage{amssymb}  
\usepackage{graphicx}
\usepackage{booktabs}
\usepackage{caption}
\usepackage{subcaption}
\usepackage{hyperref}
\usepackage{diagbox}
\usepackage{multirow}
\usepackage{bm}

\title{\LARGE \bf
Exploiting Emotional Dependencies with Graph Convolutional Networks for Facial Expression Recognition
}

\author{\parbox{16cm}{\centering
    {\large Panagiotis Antoniadis, Panagiotis Paraskevas Filntisis, Petros Maragos}\\
    {\normalsize
    School of E.C.E., National Technical University of Athens, Greece\\}}
    \thanks{This work was not supported by any organization.}
}

\begin{document}

\ifFGfinal
\thispagestyle{empty}
\pagestyle{empty}
\else
\author{Anonymous FG2021 submission\\ Paper ID \FGPaperID \\}
\pagestyle{plain}
\fi
\maketitle

\begin{abstract}
Over the past few years, deep learning methods have shown remarkable results in many face-related tasks including automatic facial expression recognition (FER) in-the-wild. Meanwhile, numerous models describing the human emotional states have been proposed by the psychology community. However, we have no clear evidence as to which representation is more appropriate and the majority of FER systems use either the categorical or the dimensional model of affect. Inspired by recent work in multi-label classification, this paper proposes a novel multi-task learning (MTL) framework that exploits the dependencies between these two models using a Graph Convolutional Network (GCN) to recognize facial expressions in-the-wild. Specifically, a shared feature representation is learned for both discrete and continuous recognition in a MTL setting. Moreover, the facial expression classifiers and the valence-arousal regressors are learned through a GCN that explicitly captures the dependencies between them. To evaluate the performance of our method under real-world conditions we perform extensive experiments on the AffectNet and Aff-Wild2 datasets. The results of our experiments show that our method is capable of improving the performance across different datasets and backbone architectures. Finally, we  also surpass the previous state-of-the-art methods on the categorical model of AffectNet.
\end{abstract}

\section{INTRODUCTION}
\label{sec:introduction}

Facial expressions are one of the most powerful nonverbal ways for human beings to convey their emotional state \cite{darwin2015expression}. Facial expression recognition (FER) has been a topic of study for decades due to its potential applications in various fields including human-computer interaction, digital entertainment, advertisement, health care and intelligent robot systems \cite{corneanu2016survey}, \cite{blom2014towards}, \cite{mcduff2014predicting}, \cite{muhammad2017facial}, \cite{filntisis2019fusing}. However, recognizing facial expressions in the wild is still very challenging due to variations, occlusions and the ambiguity of human emotion \cite{li2020deep}, \cite{valstar2011first}.

While the cultural and ethnic background of a person can affect his expressive style, Ekman indicated that humans perceive certain basic emotions in the same way regardless of their culture \cite{ekman1971constants}, \cite{ekman1994strong}. These six universal facial expressions (happiness, sadness, surprise, fear, disgust and anger) constitute the categorical model. Contempt was subsequently added as one of the basic emotions \cite{matsumoto1992more}. Due to its direct and intuitive definition of facial expressions, the categorical model is used in the majority of FER algorithms (\cite{cai2018island}, \cite{acharya2018covariance}, \cite{kervadec2018cake}, \cite{hu2017learning}, etc) and large-scale databases (AffectNet \cite{mollahosseini2017affectnet}, RAF-DB \cite{li2018reliable}, SFEW \cite{dhall2011static}, FER-2013 \cite{goodfellow2013challenges}, EmotionNet \cite{fabian2016emotionet}, etc). However, the subjectivity and ambiguity of restricting human emotion to discrete categories result in large intra-class variations and small inter-class differences.

Recently, the dimensional model proposed by Russell \cite{russell1980circumplex} has gained a lot of attention where emotion is described using a set of 2 latent dimensions that are valence (how pleasant or unpleasant a feeling is) and arousal (how likely is the person to take action under the emotional state). Another dimension called dominance is used sometimes to know whether the person is controlling the situation or not. Since a continuous representation can distinguish between subtly different displays of affect and encode small changes in the intensity, some recent algorithms \cite{nicolaou2011continuous}, \cite{chang2017fatauva}, \cite{kollias2019expression} and databases (Aff-Wild \cite{zafeiriou2017aff} and AFEW-VA \cite{kossaifi2017afew}) have utilized the dimensional model for uncontrolled FER. Even so, predicting a 2-dimensional continuous value instead of a category increases a lot the task complexity and lacks intuitiveness.

\begin{figure}[t]
    \centering
    \includegraphics[width=0.4\textwidth]{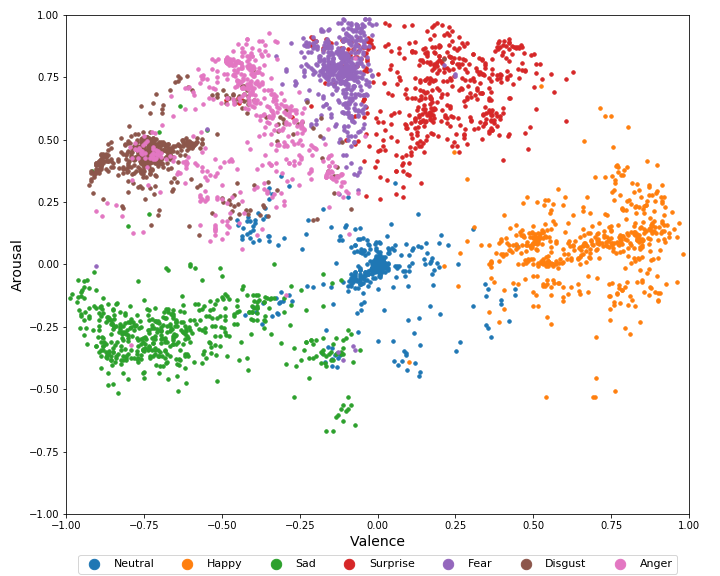}
    \caption{Distribution of the basic expressions in the VA space using the validation set of AffectNet that illustrates the emotional dependencies between the categorical and the dimensional model. The basic emotions are located around the neutral emotion that appears when valence and arousal are close to zero.}
    \label{fig:va_space}
\end{figure}

Altogether, the categorical and the dimensional model have their respective benefits and drawbacks \cite{kervadec2018cake}. Therefore, recent studies try to leverage both representations, along with Action Units (AU) detection, through multi-task learning (MTL) \cite{chang2017fatauva}, \cite{kollias2019expression}, \cite{xiaohua2019two}. However, the strong dependence between the categorical and the dimensional model is not fully exploited when they only share a common feature representation. Fig. \ref{fig:va_space} illustrates this relation using the validation set of AffectNet.

In multi-label classification, there has been a lot of research on how to properly capture and explore the correlation between labels \cite{li2014multi}, \cite{li2016conditional}. In \cite{chen2019multi}, Chen proposed ML-GCN model for multi-label image recognition that explicitly learns the labels correlation by generating the object classifiers via a Graph Convolutional Network (GCN) \cite{kipf2016semi}. Inspired by this architecture, we propose Emotion-GCN a novel GCN based MTL framework for FER in the wild. The main idea in this paper is to to generate dependent expression classifiers and valence-arousal (VA) regressors though a GCN based mapping function instead of learning them as separate parameter vectors. The generated vectors are then applied to an extracted image representation to enable end-to-end training. Hence, the dependence between the categorical and the dimensional emotion models is explicitly captured through both a shared feature representation and the dependent classifiers and regressors. Experiments on the AffectNet and Aff-Wild2 datasets indicate that Emotion-GCN increases the performance across different datasets and backbone networks managing to achieve state-of-the-art results on the categorical model of AffectNet.

The rest of this paper is organized as follows. Section \ref{sec:related_work} presents the recent work on FER systems and MTL. Section \ref{sec:proposed_method} describes the deep learning method that this paper proposes. Section \ref{sec:experiments} discusses the experimental results, providing a clear view of the accuracy improvements introduced by our method. This section also includes a description of the data used during the experiments. Finally, Section \ref{sec:conclusion} summarizes the key aspects of our work and concludes the paper.

\section{RELATED WORK}
\label{sec:related_work}

Image-based FER has been extensively studied for many years. Typically, a FER system consists of three stages: face detection, feature extraction and classification. Traditional approaches tend to conduct FER by using handcrafted features, such as Local Binary Patterns \cite{shan2009facial}, Gabor wavelets \cite{liu2002gabor}, \cite{moore2011local} and Histogram of Oriented Gradients \cite{carcagni2015facial}. While there is a lot of intuition behind these features and their performance on several lab-controlled databases is impressive, they lack generalizability and sufficient learning capacity \cite{li2020deep}. 

Later, many in-the-wild facial expression databases were developed that enabled the research of FER in more challenging environments. Deep Convolutional Neural Networks (CNNs) have achieved promising recognition performance by learning powerful high-level features \cite{zeng2018facial}, \cite{ding2017facenet2expnet}, \cite{yu2015image}, \cite{kim2015hierarchical}, \cite{hasani2020breg}, \cite{ding2020occlusion}. In \cite{li2018occlusion} and \cite{wang2020region} region-based attention networks were designed for pose and occlusion aware FER, where the regions are either cropped from landmark points or fixed positions. Facial Motion Prior Networks were proposed in \cite{chen2019facial} that generate a facial mask to focus on facial muscle moving regions. In \cite{hayale2019facial} deep Siamese neural networks equipped with a supervised loss function were used to reduce the high intra-class variation of the task. In \cite{georgescu2019local} deep and handcrafted features were combined and a local learning framework was used at test time based on SVMs. Recently, Self-Cure network \cite{wang2020suppressing} was proposed that suppresses the uncertainties caused by ambiguous expressions and the suggestiveness of the annotators.

MTL was first explored in \cite{caruana1997multitask} based on the idea that learning complementary tasks in parallel while using a shared representation improves the generalization performance. Since then, MTL has been used in many areas of computer vision, such as classification and detection \cite{sermanet2013overfeat} or geometry and regression tasks \cite{eigen2015predicting}. In the facial analysis domain, a multi-purpose algorithm for seven face-related tasks was proposed in \cite{ranjan2017all}. FATAUVA-Net \cite{chang2017fatauva} performs sequential facial attribute recognition, AU detection, and VA estimation on videos. In \cite{kollias2019expression} a holistic framework was proposed that jointly learns three facial behavior tasks (recognition of basic emotions, VA estimation and AUs detection) and two simple strategies were used for coupling the tasks during training. Closer to our work, in \cite{xiaohua2019two} a two-level attention with a two-stage MTL (2Att-2Mt) framework was proposed for facial emotion estimation on static images. However, the work was focused mainly on the estimation of VA and the dependencies between the emotion representations was not further explored. 

Apart from MTL, research on the dependencies between the categorical and the dimensional emotion representations is limited. Recently, in CAKE \cite{kervadec2018cake} the link between the two representations was explored and a 3-dimensional representation of emotion learned in a multi-domain fashion was proposed.

\section{OUR PROPOSED METHOD}
\label{sec:proposed_method}

\begin{figure*}[t]
    \centering
    \includegraphics[width=\textwidth]{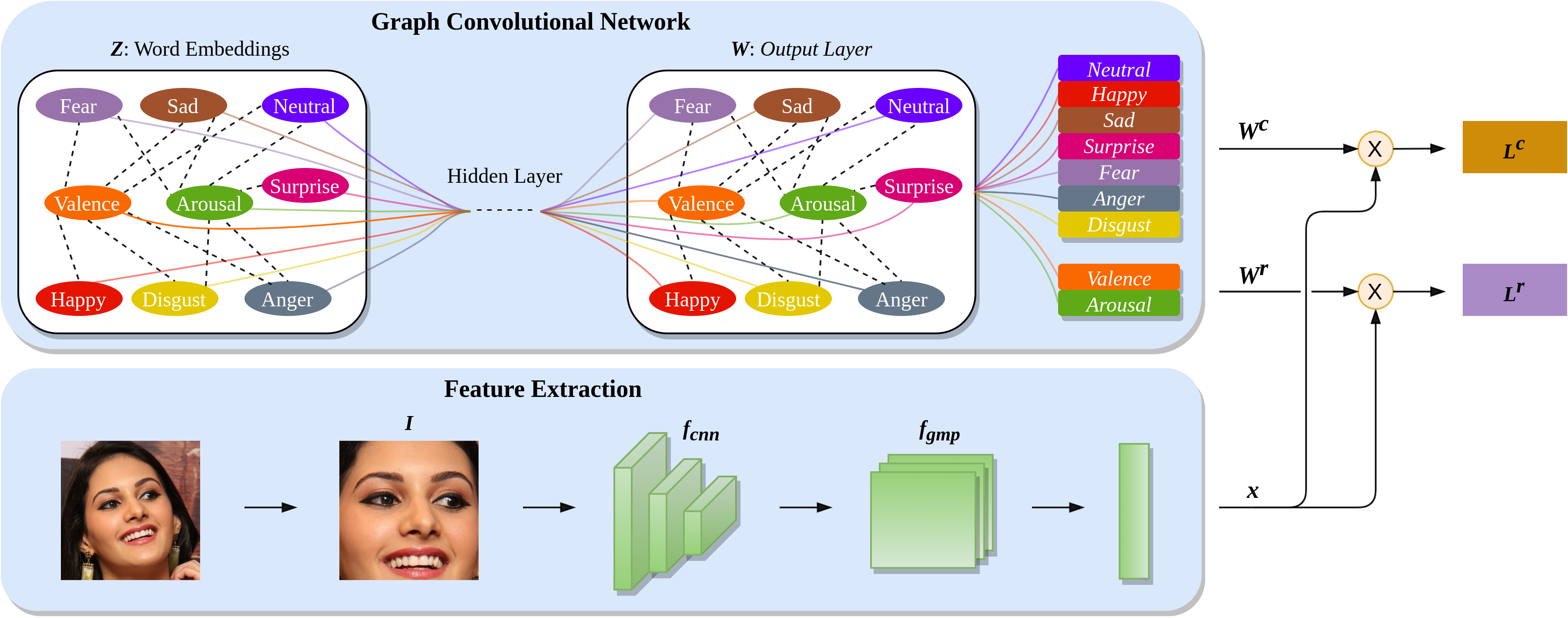}
    \caption{Overall architecture of our Emotion-GCN model for FER in the wild. Our graph contains the seven expression labels and the two VA dimensions that are connected to each other based on the adjacency matrix $\mathbf{A}$. Stacked GCNs are learned over the graph to map the word embeddings of the nodes $\mathbf{Z}$ into a set of dependent classifiers and regressors, $\mathbf{W^c}$ and $\mathbf{W^r}$ respectively. These vectors are then applied to an image representation $\mathbf{x}$ extracted from the input image $\mathbf{I}$ via a DenseNet $\mathbf{f_{cnn}}$ followed by a global max-pooling function $\mathbf{f_{gmp}}$. The whole network is trained end-to-end for both basic expression classification ($\mathbf{L^c}$) and VA regression ($\mathbf{L^r}$).}
    \label{fig:overview}
\end{figure*}

In this section the architecture of the proposed network, the dependent classifiers and regressors and the MTL setting for recognition are introduced. The overall architecture of the proposed Emotion-GCN is shown in Fig. \ref{fig:overview}.

\subsection{Feature extraction}

Given an input image $\mathbf{I}$ of size $227 \times 227$ pixels, we obtain $1024 \times 7 \times 7$ feature maps using a Dense Convolutional Network (DenseNet) \cite{huang2017densely}. Each layer in DenseNet obtains additional inputs from all preceding layers and passes on its own feature-maps to all subsequent layers to ensure maximum information flow between layers. Then, a global max-pooling function is applied to get the feature vector $\mathbf{x}$ of the image:
\begin{equation}
    \mathbf{x} = f_{gmp}(f_{cnn}(\mathbf{I})) \in \mathbb{R}^{D}
\end{equation}
where $f_{cnn}$ corresponds to the convolution layers of the DenseNet, $f_{gmp}$ to the the global max-pooling function and $D = 1024$.

\subsection{Dependent classifiers and regressors}

Given the feature vector $\mathbf{x}$ of the image, we want to learn one classifier for each facial expression (classification task) and one regressor for each dimension in the VA space (regression task). Each classifier acts as a weight vector that is multiplied with the feature $\mathbf{x}$ and then passes though a softmax activation function assigning a probability score to its emotion category. In parallel, each regressor follows the same procedure without passing through an activation function since its output is a  predicted continuous value for either valence or arousal. We denote the expression classifiers by $\mathbf{W^{c}} \in \mathbb{R}^{7 \times D}$ and the VA regressors by $\mathbf{W^{r}} \in \mathbb{R}^{2 \times D}$ where each row of the matrices corresponds to a classifier and a regressor respectively. Traditionally, these vectors are optimized as individual parameters of the deep learning network. To capture the correlation between the categorical and the dimensional model, we propose to learn these vectors using a GCN that maps their word embeddings to dependent classifiers and regressors retaining the shared information between the two tasks.

\subsubsection{GCN Overview}

Generally, the goal of a GCN is to learn a function $f$ on a graph $G=(V, E)$ that takes as input a feature description for each node of the graph $\mathbf{H^l} \in \mathbb{R}^{n \times d}$ ($n = |V|$) and a correlation matrix $\mathbf{A} \in \mathbb{R}^{n \times n}$ and produces new node-level features $\mathbf{H^{l+1}} \in \mathbb{R}^{n \times d'}$. The update rule is formulated as follows:
\begin{equation}
    \mathbf{H^{l+1}} = h(\mathbf{\hat{A}} \mathbf{H^l} \mathbf{W^l})
    \label{eqn:gcn_update}
\end{equation}
where $\mathbf{W^l} \in \mathbb{R}^{d \times d'}$ is the weight matrix for the $l$-th GCN layer, $\mathbf{\hat{A}}$ is the normalized version of matrix $\mathbf{A}$ such that all rows sum to one and $h(\cdot)$ is LeakyRELU \cite{maas2013rectifier}. For more information on GCN we refer readers to \cite{kipf2016semi}.

In our case, the nodes of the graph correspond to the seven expression labels and the two VA dimensions i.e. V = \{Neutral, Happy, Sad, Surprise, Fear, Disgust, Anger, Valence, Arousal\}. Hence, the input is $\mathbf{Z} \in \mathbb{R}^{9 \times d}$ that contains the word embedding of each node ($d$ is the dimensionality of the embeddings). Each GCN layer $l$ takes the node representations from the previous layer $\mathbf{H^l}$ as inputs and outputs new node representations  $\mathbf{H^{l+1}}$ using (\ref{eqn:gcn_update}). Finally, the output representation of the last layer $\mathbf{W} \in \mathbb{R} ^ {9 \times D}$ contains the dependent classifiers and regressors. The first seven rows of the output matrix $\mathbf{W}$ constitute the classification part $\mathbf{W^{c}}$ and the rest two the regression part $\mathbf{W^{r}}$.

\subsubsection{Adjacency Matrix Design}

\begin{figure}[t]
    \centering
    \includegraphics[width=0.4\textwidth]{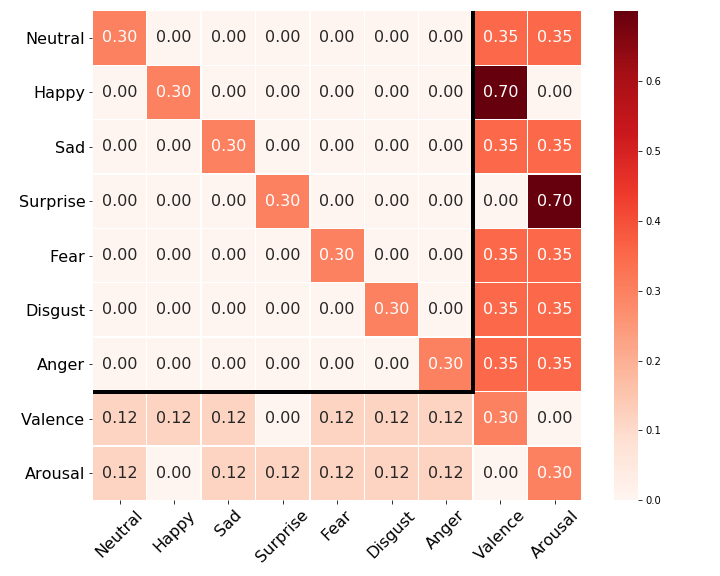}
    \caption{Adjacency matrix of our Emotion-GCN.}
    \label{fig:adj_matrix}
\end{figure}

According to the update rule of a GCN (\ref{eqn:gcn_update}) the feature of a node in the graph is the weighted sum of its own feature and the adjacent nodes’ features. Since the purpose behind the use of a GCN is to exploit the dependencies between the categorical and the dimensional model, we should design the adjacency matrix $\mathbf{A}$ to this direction. As before, we assume that the first seven rows of $\mathbf{A}$ correspond to the basic expressions and the last two are the continuous dimensions i.e. valence and arousal. Since we deal with a multi-task and not a multi-label recognition problem, we are interested only in the correlation between the categorical and the dimensional model. Hence, we set the other pairs of $\mathbf{A}$ to zero except for the diagonal to allow self-loops. Also, we take the absolute value of the correlation to ignore its type (positive or negative) and focus on its amplitude. The correlation matrix $\mathbf{A} \in \mathbb{R}^{9 \times 9}$ can be written as:

\begin{equation}
    A_{ij} = \begin{cases}
    1, &\ \text{if} \ \ i = j \\
    |c_{ij}|, &\ \text{if} \ \ i \in \textit{Cat} \ \land \ j \in \textit{Dim} \\
    |c_{ij}|, &\ \text{if} \ \ j \in \textit{Cat} \ \land \ i \in \textit{Dim} \\
    0, &\ \text{else}
    \end{cases}
\end{equation}
where \textit{Cat} and \textit{Dim} are the set of indices of the categorical and the dimensional labels respectively. As a correlation metric, we use the Spearman's rank correlation coefficient \cite{spearman1961proof} that for two variables $\mathbf{X} = \{x_1, ..., x_N\}$ and $\mathbf{Y} = \{y_1, ..., y_N\}$ is defined as:

\begin{equation}
    c_{xy} = \frac{\sum_{k=1}^N x_{k,r} y_{k,r}}{\sqrt{\sum_{k=1}^N x_{k,r}^2 \sum_{k=1}^N y_{k,r}^2}}
\end{equation}
where $x_{k, r}$ and $y_{k, r}$ are the rank transformation of the initial values $x_k$ and $y_k$. Following the ideas in ML-GCN \cite{chen2019multi}, we use a threshold $\tau$ to filter the noisy edges as follows:

\begin{equation}
    A'_{ij} = \begin{cases}
    1, &\ \text{if} \ A_{ij} \geq \tau \\
    0, &\ \text{if} \ A_{ij} < \tau
    \end{cases}
\end{equation}
where $\tau=0.1$ to enable the propagation of information between weakly correlated nodes. As shown in \cite{li2018deeper}, the learned features of each node may be over-smoothed and become indistinguishable when a binary correlation matrix is used. To alleviate the over-smoothing problem, we apply the re-weighted scheme of ML-GCN that is defined as: 

\begin{equation}
    A^{''}_{ij} = \begin{cases}
    (p/\sum_{\substack{j=1 \\ i \neq j}}^{9} A'_{ij}) \times A'_{ij}, &\ \text{if} \ i \neq j \\
    1-p, &\ \text{if} \ i = j
    \end{cases}
    \label{eqn:adj_type}
\end{equation}
where the variable $p$ determines the weights assigned to a node itself and its adjacent nodes. We set $p = 0.7$ to increase the influence of the neighborhood nodes. Fig. \ref{fig:adj_matrix} shows the output adjacency matrix $\mathbf{A^{''}}$ of Emotion-GCN using the training set of AffectNet. As we expected, pairs like happy-valence and surprise-arousal are strongly connected since the presence of the one usually denotes the presence of the other (Fig. \ref{fig:va_space}).

\subsection{Multi-task learning setting}

After we generate the dependent matrices $\mathbf{W^{c}}$, $\mathbf{W^{r}}$ and the feature vector $\mathbf{x}$, we perform FER simultaneously on the categorical and the dimensional model. For the classification task, the predicted scores are computed as:

\begin{equation}
    \hat{y}_i = \frac{e^{\mathbf{w_i^{c}} \cdot \mathbf{x}}}{\sum_{k=1}^7 e^{\mathbf{w_k^{c}} \cdot \mathbf{x}}}
\end{equation}
where $\hat{y}_i$ is the probability of emotion $i$ and $\mathbf{w_i^{c}}$ is the classifier of emotion $i$. The network is trained using a weighted version of the traditional categorical cross entropy loss $\mathbf{L^{c}}$ since the dataset is highly imbalanced \cite{mollahosseini2017affectnet}. In other words, the network is penalized more for misclassifying samples from under-represented classes than from well-represented classes as follows:

\begin{equation}
    L^{c} = - \sum_{i=1}^7 w_i \ y_i \log (\hat{y}_i)
    \label{eqn:weighted_ce}
\end{equation}

\begin{equation}
    w_i = \frac{f_i}{f_{min}}
    \label{eqn:weight}
\end{equation}
where $y_i = 1$ if class $i$ is the ground truth expression, $f_i$ is the number of samples of the $i$-th class and $f_{min}$ is the number of samples in the most under-represented class i.e. Disgust. For the regression task, the predicted values are computed as:
\begin{equation}
    \hat{p}_i = \mathbf{w^{r}_i} \cdot \mathbf{x}
\end{equation}
where $\hat{p}_1, \hat{p}_2$ are the predicted values of valence and arousal respectively and $ \mathbf{w^{r}_1}, \mathbf{w^{r}_2}$ correspond to the VA regressors. Most studies in the literature use the Mean Squared Error (MSE) as a loss function to train regression models. However, more and more works on emotion recognition use a Correlation-based loss function to measure the agreement between the true emotion dimension and the predicted emotion degree \cite{kollias2019expression}, \cite{han2017hard}. The Concordance Correlation Coefficient (CCC) is often used since it takes the bias into Pearson’s correlation coefficient \cite{atmaja2020evaluation}. It is defined as:

\begin{equation}
    \rho_c = \frac{2 s_{xy}}{s_x^2 + s_y^2 + (\bar{x} - \bar{y})^2}
\end{equation}
where $s_x$ and $s_y$ denote the variance of the predicted and ground truth values respectively, $\bar{x}$ and $\bar{y}$ are the corresponding mean values and $s_{xy}$ is the respective covariance value. The range of CCC is from -1 (perfect disagreement) to 1 (perfect agreement). Hence, in our case we define $\mathbf{L^{r}}$ as:

\begin{equation}
    L^{r} = 1 - \frac{\rho_v + \rho_a}{2}
    \label{eqn:ccc}
\end{equation}
where $\rho_v$ and $\rho_a$ are the respective CCC of valence and arousal. The overall loss function of our network training is defined as $L = L^{c} + L^{r}$.

\section{EXPERIMENTS}
\label{sec:experiments}

In this section, we present the experimental evaluation of the proposed Emotion-GCN model (Fig. \ref{fig:overview}). First, we briefly describe the facial expression datasets that we use and the details of our implementations. Then, an ablation study of our models follows and finally a comparison with the current state-of-the-art methods.

\subsection{Datasets}

\subsubsection{AffectNet} AffectNet is by far the largest database of facial expressions that provides both categorical and VA annotations. It contains more than one million facial images collected from the Internet by querying three major search engines using 1,250 emotion-related keywords in six different languages \cite{mollahosseini2017affectnet}. About half of the retrieved images are manually annotated for the presence of seven discrete facial expressions (categorical model) and the intensity of valence and arousal (dimensional model). It is a very challenging database as it contains images of people from different races and ethnic groups as well as high variety in the background, lighting, pose, point of view, etc. To be consistent with the majority of past studies \cite{zeng2018facial}, \cite{li2018occlusion}, \cite{chen2019facial}, \cite{kervadec2018cake}, \cite{ding2020occlusion}, \cite{georgescu2019local}, \cite{hayale2019facial}, we exclude the emotion of contempt and train our models on the images with neutral and the 6 basic emotions (approximately 280,000 training samples). Since the test set of AffectNet is not publicly available, we evaluate our approaches on the validation set that contains 500 images for each emotion. The mean class accuracy is used as the evaluation metric for the classification task because the validation set is balanced and the CCC for the regression task.

\subsubsection{Aff-Wild2} Aff-Wild2 is the first ever database annotated for valence-arousal estimation, action unit detection and basic expression classification \cite{kollias2019expression}. It consists of 548 videos collected from YouTube and shows both subtle and extreme human behaviours in real-world settings. We trained and evaluated our models on a subset of the database that contains only the frames with both categorical and VA annotations, as required by our method. As proposed by the authors of Aff-Wild2 a weighted average between the F1 score and mean class accuracy is used as the evaluation metric for the classification task ($0.67 \times \textit{F1} + 0.33 \times \textit{Acc}$).

\subsection{Implementation details}

Before training the network, the face region of each image is cropped using the provided bounding box by the database and scaled to $227 \times 227$ pixels. Also, we perform face alignment based on the position of the eyes to obtain a normalized representation of each face. Specifically, we compute the location of the eyes by taking the mean value of the six detected landmarks in each eye. Then, we rotate the image by $\theta$ that is defined as $\theta = tan^{-1}\left(\frac{r_y - l_y}{r_x - l_x}\right)$, where $(r_x, \ r_y)$ and $(l_x, \ l_y)$ are the coordinates of the right and the left eye respectively. To augment the data, six types of augmentation techniques are used (flip, rotation and changes in brightness, contrast, hue and saturation) \cite{hayale2019facial}. The GCN module consists of two layers defined with output dimensionality of 512 and 1024 respectively. For the word representations, we use 300-dimensional GloVe \cite{pennington2014glove} trained on the Wikipedia dataset. The networks are trained for 10 epochs using a batch size of 35 and a learning rate of 0.001. Stochastic Gradient Descent is adopted as the optimization algorithm with a momentum of 0.9 and PyTorch \cite{paszke2017automatic} is used as our deep learning framework.

\subsection{Ablation Study}

First, we investigate the performance of four different networks on the categorical and the dimensional model of affect to present the improvements that Emotion-GCN introduces.

\subsubsection{Categorical Model} 

\begin{table}[t]
  \caption{Performance of our models on the categorical model of AffectNet and Aff-Wild2.}
  \label{table:results_ours_cat}
  \begin{center}
  \begin{tabular}{lcc}
    \textbf{Method} & \textbf{AffectNet} & \textbf{Aff-Wild2} \\
    \toprule
    Single-task            & 64.37 & 45.06 \\
    Multi-task + MSE       & 64.8 & 43.1 \\
    Multi-task + CCC       & 65.69 & 43.33\\
    Emotion-GCN            & \textbf{66.46} & \textbf{48.92}\\
  \end{tabular}
  \end{center}
\end{table}

\begin{figure*}[t]
     \centering
     \begin{subfigure}[b]{0.24\textwidth}
         \centering
         \includegraphics[width=\textwidth]{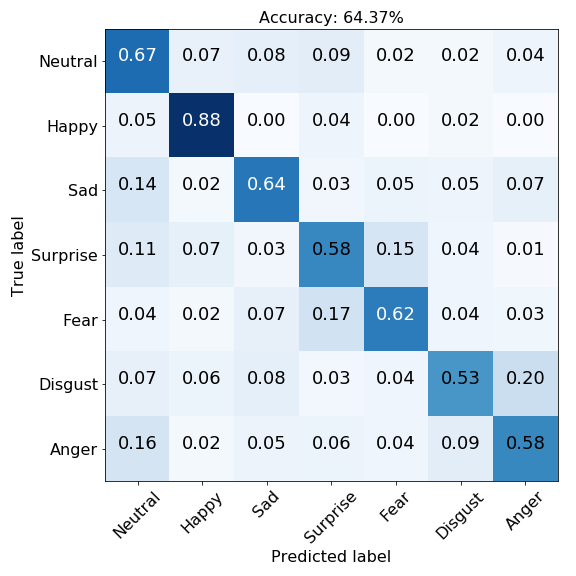}
         \caption{Single-task}
         \label{fig:cm_single}
     \end{subfigure}
     \hfill
     \begin{subfigure}[b]{0.24\textwidth}
         \centering
         \includegraphics[width=\textwidth]{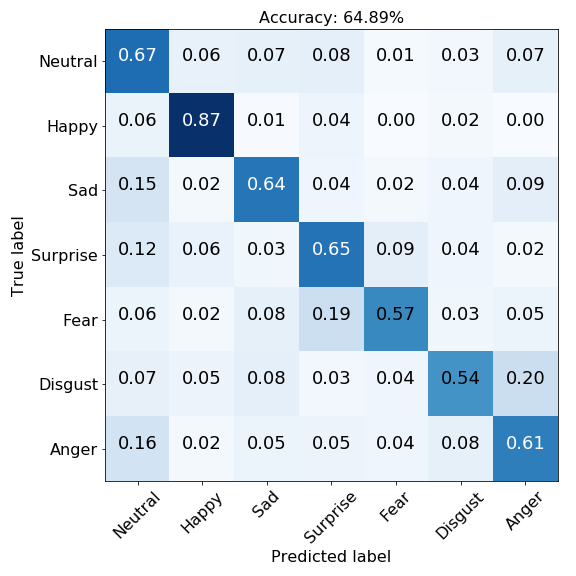}
         \caption{Multi-task + MSE}
         \label{fig:cm_multi_mse}
    \end{subfigure}
     \hfill
     \begin{subfigure}[b]{0.24\textwidth}
         \centering
         \includegraphics[width=\textwidth]{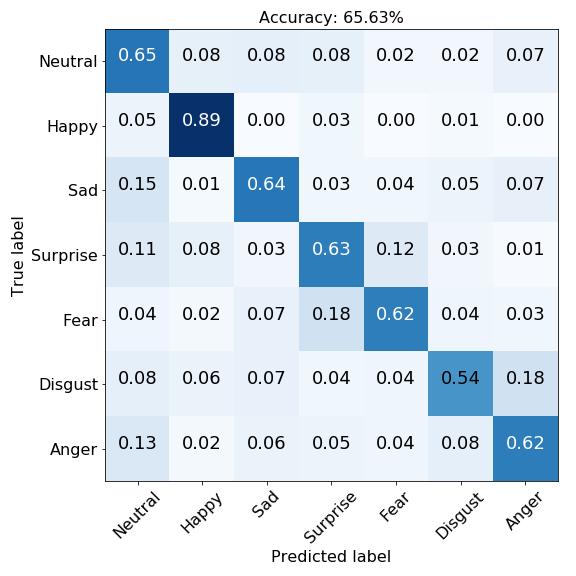}
         \caption{Multi-task + CCC}
         \label{fig:cm_multi_ccc}
     \end{subfigure}
     \hfill
     \begin{subfigure}[b]{0.24\textwidth}
         \centering
         \includegraphics[width=\textwidth]{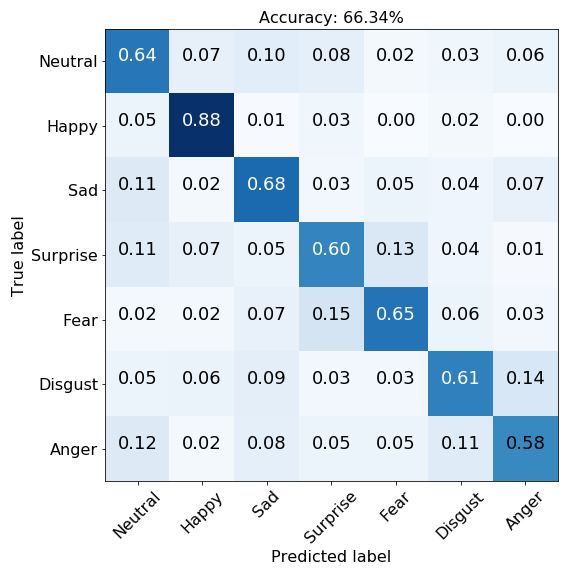}
         \caption{Emotion-GCN}
         \label{fig:cm_multi_gcn}
    \end{subfigure}
    \caption{Confusion matrices of our models on the validation set of AffectNet.}
    \label{fig:cm}
\end{figure*}

We trained (i) a single-task network for discrete FER using the weighted CE loss of (\ref{eqn:weighted_ce}). Then, two multi-task networks were trained for discrete and continuous FER using a weighted CE loss for the classification task and a (ii) MSE or (iii) CCC loss (\ref{eqn:ccc}) for the regression task. Finally, we trained (iv) the proposed Emotion-GCN model that generates dependent classifiers and regressors using a 2-layer GCN as presented in Fig. \ref{fig:overview}. Table \ref{table:results_ours_cat} shows the results of our experiments. In AffectNet we can see that learning to predict the VA values as an additional task in a MTL framework increases the accuracy by \textbf{1.32\%} since the shared representation improves the generalization of the network. Also, in agreement with recent studies, using the CCC loss for the regression task increases the classification accuracy of the model verifying that in MTL a correlation-based regression loss performs better. Finally, Emotion-GCN achieves a total accuracy of \textbf{66.46\%}, since the dependent classifiers manage to effectively capture the dependencies between the two emotion representations. The confusion matrices for these models are shown in Fig. \ref{fig:cm}. It can be seen that our proposed method increases the accuracy for most classes while the single-task network performs better only in the neutral class. Similarly, in Aff-Wild2 the total evaluation metric increases by \textbf{3.86\%} indicating that the benefits of Emotion-GCN generalize in more datasets. To investigate  how different choices for the parameters of the GCN affect the performance, we perform additional experiments in Table \ref{table:results_parameters}. We observe that the number of GCN layers ($L$) is the most crucial parameter since using only one layer decreases the accuracy a lot.

\begin{table}[t]
  \caption{Classification accuracy of Emotion-GCN on the categorical model of AffectNet using different values for $\tau$, $p$ and $L$ (number of GCN layers). In the first table $p=0.7$ and in the second table $\tau=0.1$.} 
  \label{table:results_parameters}
  \begin{center}
  \begin{tabular}{ccccc}
    \diagbox{$L$}{$\tau$} & 0 & 0.05 & 0.1 & 0.15 \\
    \toprule
    1 & 53.09 & 43.85 &  53.71	& 54.94\\
    2 & 65.29 & 65.31 & \textbf{66.46} & 65.69\\
    3 &	65.77 & 64.94 & 64.74	& 65.74\\
  \end{tabular}
  \begin{tabular}{cccccccc}
    \diagbox{$L$}{$p$} & 0.2 & 0.3 & 0.4 & 0.5 & 0.6 & 0.7 & 0.8 \\
    \toprule
    1 & 49.6 & 50.94 & 52.57 & 52.66 & 52.86 & 53.71 & 59.31 \\
    2 & 65.71 & 66.14 & 65.29 & 66.29 & 66.09 & \textbf{66.46} & 65.06 \\
    3 & 65.29 & 65.54 & 65.89 & 66.09 & 65.54 & 64.74 & 65.83 \\
  \end{tabular}
  \end{center}
\end{table}

\subsubsection{Dimensional Model} 

\begin{table}[t]
  \caption{Performance of our models on the dimensional model of AffectNet and Aff-Wild2.}
  \label{table:results_ours_dim}
  \begin{center}
  \begin{tabular}{lcccc}
    \multirow{2}{*}{\textbf{Method}} & \multicolumn{2}{c}{\textbf{AffectNet}} & \multicolumn{2}{c}{\textbf{Aff-Wild2}} \\
    & CCC-V & CCC-A & CCC-V & CCC-A \\
    \toprule
    Single-task            & 0.761 & 0.628 & 0.416 & 0.501\\
    Multi-task + MSE       & 0.752 & 0.572 & 0.435 & 0.378\\
    Multi-task + CCC       & \textbf{0.768} & \textbf{0.651} & 0.408 & 0.481\\
    Emotion-GCN            & 0.767 & 0.649 & \textbf{0.457} & \textbf{0.514}\\
  \end{tabular}
  \end{center}
\end{table}

\begin{figure*}[t]
    \centering
    \includegraphics[width=\textwidth]{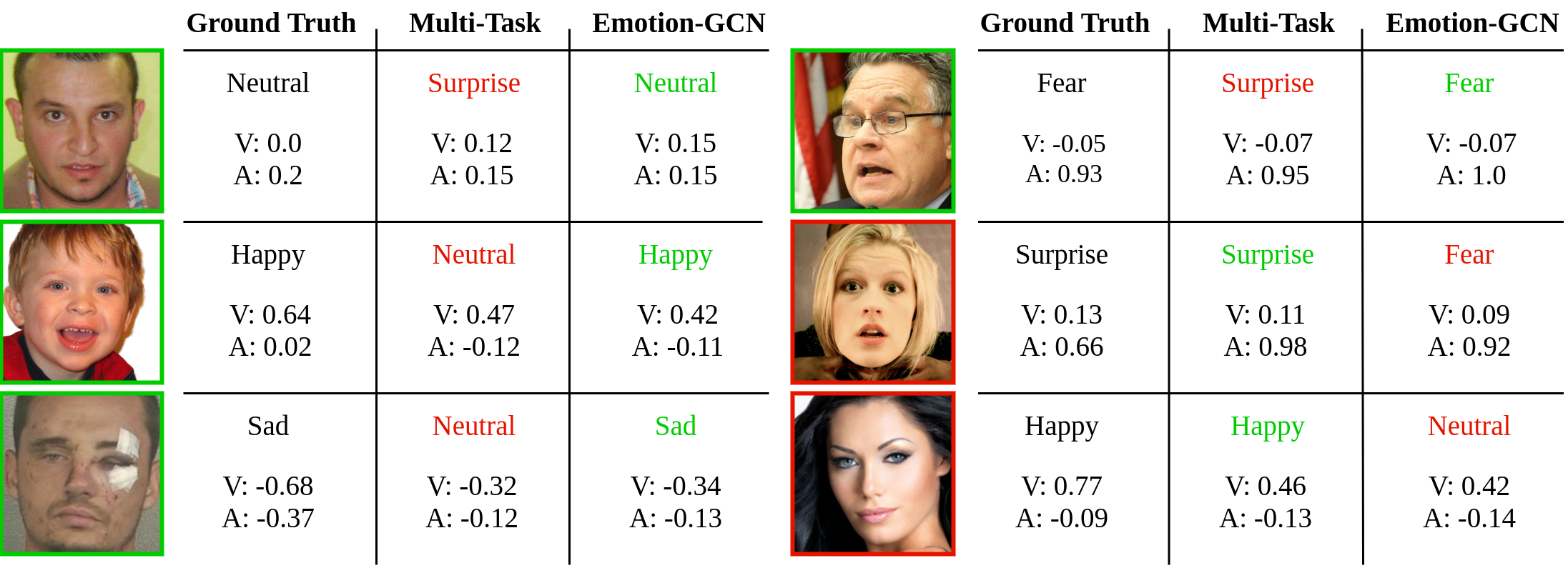}
    \caption{Predictions of our models on samples of AffectNet. The first column indicates the ground truth values. The second and the third column present the predictions of the multi-task network trained with CCC loss and our Emotion-GCN respectively. To examine which network better exploits the emotional dependencies, we selected samples where the predictions of the networks on the dimensional model are close. In the first four samples our Emotion-GCN model successfully recognizes the depicted emotion while the multi-task network fails indicating that our proposed model effectively captured the dependencies presented in the VA space (Fig. \ref{fig:va_space}). However, there are cases where our network fails (last two samples).}
    \label{fig:samples}
\end{figure*}

For the evaluation of our models on the VA space, the networks (ii), (iii) and (iv) are the same since they are trained for both discrete and continuous FER. Additionally, a single-task network for VA regression was trained using the CCC loss of (\ref{eqn:ccc}). Table \ref{table:results_ours_dim} shows the performance evaluation of our experiments on the dimensional model. In AffectNet we can see that the multi-task network with the CCC loss improves the regression performance along with the classification one verifying that both tasks benefit from the shared feature representation. Actually, we observe that the increase in the performance of arousal prediction is higher that that of valence. This is due to the fact that most emotions have positive value of arousal (Fig. \ref{fig:va_space}) and a simultaneous emotion classification provides more useful information in the task of arousal regression. Finally, our GCN based approach achieves similar performance with that of the multi-task network on the dimensional model. In Aff-Wild2, our proposed model surpasses the performance of both the single-task and multi-task networks. Overall, our method presents significant improvements in both the categorical and the dimensional model. In Fig. \ref{fig:samples} we present some positive and negative results of our Emotion-GCN model on AffectNet.

\subsection{Visualization}

\begin{figure*}[t]
     \centering
     \begin{subfigure}[b]{0.32\textwidth}
         \centering
         \includegraphics[width=\textwidth]{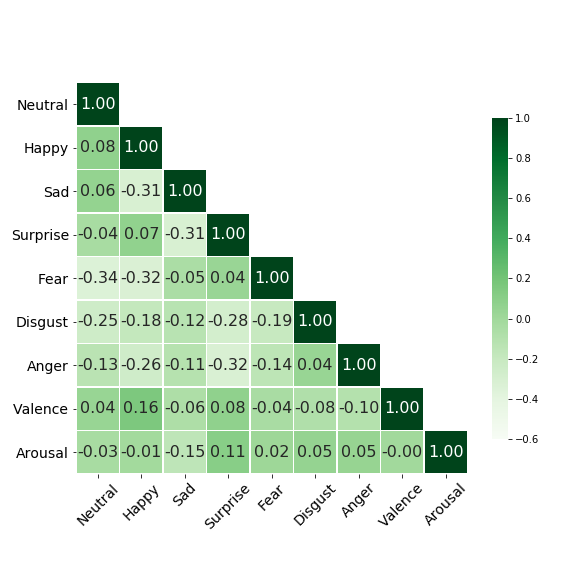}
         \caption{Single-task}
         \label{fig:similarity_single}
     \end{subfigure}
     \hfill
     \begin{subfigure}[b]{0.32\textwidth}
         \centering
         \includegraphics[width=\textwidth]{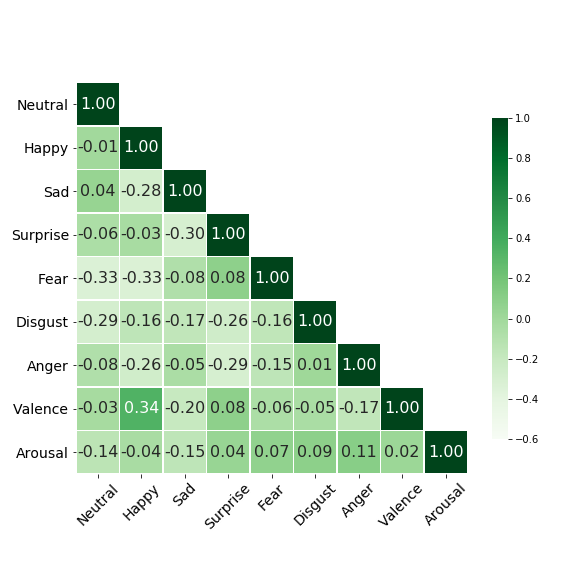}
         \caption{Multi-task + CCC}
         \label{fig:similarity_multi}
     \end{subfigure}
     \hfill
     \begin{subfigure}[b]{0.32\textwidth}
         \centering
         \includegraphics[width=\textwidth]{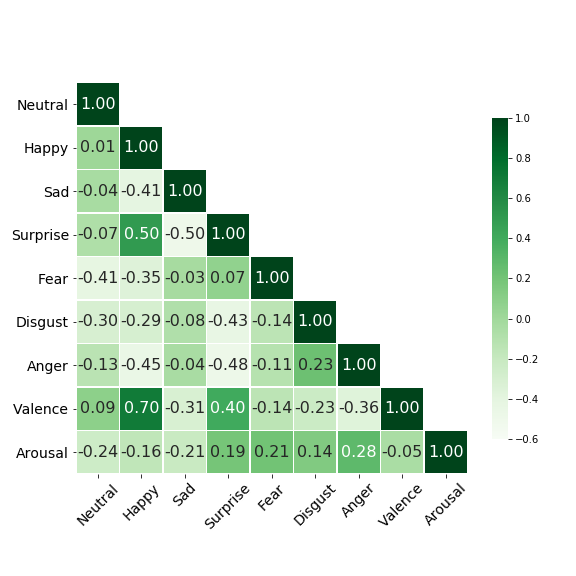}
         \caption{Emotion-GCN}
         \label{fig:similarity_gcn}
     \end{subfigure}
    \caption{Visualization of the cosine similarity between the learned classifiers and regressors by our models on AffectNet.}
    \label{fig:similarity}
\end{figure*}

To further analyze the effectiveness of our approach, we investigate the similarity between the dependent classifiers and regressors. In Fig. \ref{fig:similarity} the cosine similarity of the learned vectors on AffectNet by our single-task networks, our best multi-task network and our Emotion-GCN are presented. As we can see, learning a shared representation for both tasks through MTL slightly increases the similarity between the expression classifiers and the VA regressors. Our proposed method increases their similarity even more in consistence with the dependencies presented in Fig. \ref{fig:va_space}. Specifically, the regressor of valence comes closer to the classifier of happy and the regressor of arousal closer to the classifiers of anger, disgust, fear and surprise. These emotions appear in regions of VA space where the values of valence or arousal are high. Therefore, the proposed network has successfully captured the dependence between the categorical and the dimensional emotion representation.

We also observe an increase in the similarity between the pairs valence-surprise and happy-surprise while their respective nodes are not adjacent in the graph (Fig. \ref{fig:adj_matrix}). These dependencies are successfully captured by our network since in a 2-layer GCN a node incorporates information from a 2-hop neighborhood \cite{derr2018signed}. To further examine whether the dependence between the emotions of happiness and surprise is reasonable, we compute their co-occurrence in a multi-label emotion dataset. The EMOTIC dataset \cite{kosti2017emotic} is a collection of images of people in unconstrained environments annotated according to their apparent emotional states. Each person is annotated for 26 discrete categories, with multiple labels assigned to each image. About 25\% of the samples labeled as happy are labeled as surprise too that indicates that these two emotions are strongly related indeed.

\subsection{Dependence between classifiers}

Inspired by the fact that there are dependencies between the basic emotions, we trained a similar GCN based network but we enabled this time the direct propagation of information between the seven emotion classifiers. Instead of setting their correlation to zero as in Emotion-GCN, we compute the conditional probability matrix of the basic emotions like in ML-GCN. The model achieves a total accuracy of 66.23\% on the categorical model of AffectNet. Despite the intuition behind this approach, the recognition performance is slightly lower than that of our proposed method. We believe that by enabling the propagation of information between the basic emotions the dependence between the categorical and dimensional model is ignored. Also, we deal with a single-label recognition task and the possible benefits of this approach are more suitable in a multi-label recognition task.

\subsection{Comparison with the State of the Art}
In Table \ref{table:results_sota}, we compare the performance of our Emotion-GCN model with several state-of-the-art methods for FER on the categorical model of AffectNet. Regarding Aff-Wild2, we only used the subset which contains both categorical and continuous annotations, while methods in the bibliography typically report results on the whole dataset, making direct comparisons not possible. In IPA2LT \cite{zeng2018facial}, the authors designed LTNet to discover the latent truths from the human annotations and the machine annotations trained from different FER datasets. In gACNN \cite{li2018occlusion} and OADN \cite{ding2020occlusion} attention networks were proposed for occlusion aware FER. Facial Motion Prior Network in \cite{chen2019facial} generates a facial mask so as to focus on facial muscle moving regions. In \cite{georgescu2019local} deep (CNNs) and handcrafted features (BOVW) were combined and in \cite{hayale2019facial} a deep Siamese network along with a supervised loss function was used to reduce the intra-class variation of the task. Closer to our work, CAKE \cite{kervadec2018cake} proposed a 3-dimensional representation of emotion learned in a multi-domain fashion. Our Emotion-GCN model considerably outperforms these recent state-of-the-art methods, achieving an accuracy of \textbf{66.46\%}. 

In \cite{hasani2020breg} BReG-NeXt achieved state-of-the-art accuracy in 8-way classification on AffectNet (including contempt) by introducing a residual-based network architecture. To investigate the ability of Emotion-GCN to generalize over other model architectures as well, we replaced our DenseNet backbone network with BReG-NeXt using the publicly available code\footnote{\url{https://github.com/behzadhsni/BReG-NeXt}}. Since the provided code does not include the data preprocessing strategy, we followed our preprocessing and training pipeline described before (same with DenseNet), which explains the different results compared to~\cite{hasani2020breg}. As we can see in Table \ref{table:results_breg}, our proposed model outperforms the single-task and multi-task models when using BReG-NeXt as the backbone network indicating that Emotion-GCN generalizes across different model architectures as well.

\begin{table}[t]
  \caption{Comparison with state-of-the-art methods on AffectNet (7-way classification).}
  \label{table:results_sota}
  \begin{center}
  \begin{tabular}{lc}
    \textbf{Method} & \textbf{Accuracy} \\
    \toprule
    IPA2LT \cite{zeng2018facial}                         & 57.31 \\
    gACNN \cite{li2018occlusion}                         & 58.78 \\
    Facial Motion Prior Network \cite{chen2019facial}    & 61.52 \\
    CAKE \cite{kervadec2018cake}                         & 61.7 \\
    OADN \cite{ding2020occlusion}                        & 61.89 \\
    CNNs and BOVW + global SVM \cite{georgescu2019local} & 63.31 \\
    Siamese \cite{hayale2019facial}                      & 64 \\
    \textbf{Emotion-GCN (ours)}   & \textbf{66.46} \\
  \end{tabular}
  \end{center}
\end{table}

\begin{table}[t]
  \caption{Performance of Emotion-GCN using BReG-NeXt as the backbone network on AffectNet.}
  \label{table:results_breg}
  \begin{center}
  \begin{tabular}{lcc}
    \textbf{Method} & \textbf{Network} & \textbf{Accuracy} \\
    \toprule
    Single-task & BReG-NeXt                                            & 60.49\\
    Multi-task+CCC & BReG-NeXt                               & 61.14\\
    Emotion-GCN & BReG-NeXt                              & \textbf{61.94}\\
  \end{tabular}
  \end{center}
\end{table}

\section{CONCLUSION}
\label{sec:conclusion}

In this work, a novel GCN based MTL framework is proposed for in-the-wild FER. Specifically, our Emotion-GCN model learns a shared feature representation for both discrete and continuous expression recognition to exploit the dependencies between the categorical and the dimensional model of affect. To further capture these dependencies, the expression classifiers and the VA regressors are learned though a GCN that maps their word representation to dependent vectors inspired by recent work in multi-label image recognition. Experimental results on AffectNet, the largest facial expression database, have demonstrated that our Emotion-GCN outperforms the performance of the recent state-of-the-art methods for discrete FER.

\addtolength{\textheight}{0cm}   

{\small
\bibliographystyle{ieee}
\bibliography{paper}
}

\end{document}


\ifFGfinal
\thispagestyle{empty}
\pagestyle{empty}
\else
\author{Anonymous FG2021 submission\\ Paper ID \FGPaperID \\}
\pagestyle{plain}
\fi
\maketitle


\begin{abstract}
In the paper we presented Emotion-GCN, a novel MTL framework that exploits the dependencies between the categorical and the dimensional model using a GCN to recognize facial expressions in-the-wild. In subsection IV. E, we trained a network similar to Emotion-GCN but we enabled this time the direct propagation of information between the seven emotion classifiers. Here, we present some additional figures related to this network.
\end{abstract}

\subsection{Adjacency Matrix}

\begin{figure}[t]
    \centering
    \includegraphics[width=0.45\textwidth]{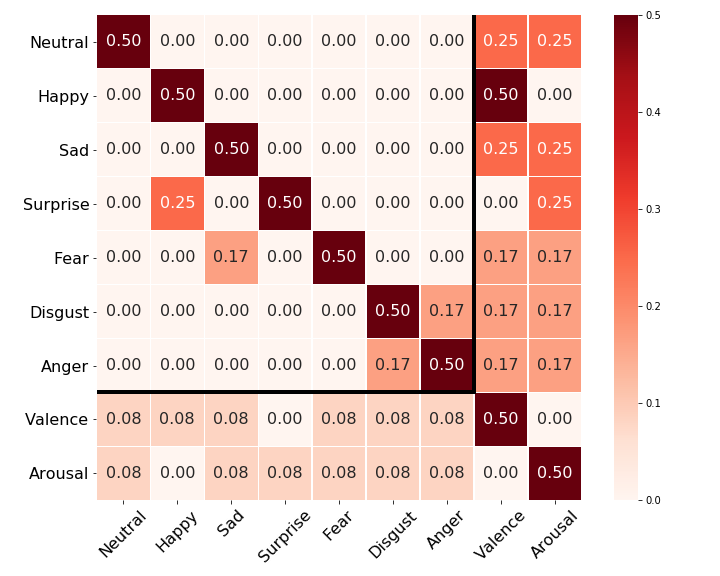}
    \caption{Adjacency matrix of a GCN that enables the propagation of information between the expression classifiers.}
    \label{fig:adj_matrix_intra}
\end{figure}

The part of the adjacency matrix that concerns the correlation between the categorical and the dimensional model is computed like in Emotion-GCN. However, to compute the correlation between the basic emotions a multi-label emotion dataset is required. We used the EMOTIC dataset \cite{kosti2017emotic} and followed the procedure of ML-GCN \cite{chen2019multi} to compute the conditional probabilities between the basic emotions. The parameters remained the same as  in Emotion-GCN model ($\tau=0.1$ and $p = 0.5$). The output adjacency matrix of the network is shown in Fig. \ref{fig:adj_matrix_intra}. As expected, some extra edges were created between the nodes of surprise-happy, fear-sad and disgust-anger enabling the propagation of information between these classifiers through training.

\subsection{Visualization}

\begin{figure}[t]
    \centering
    \includegraphics[width=0.45\textwidth]{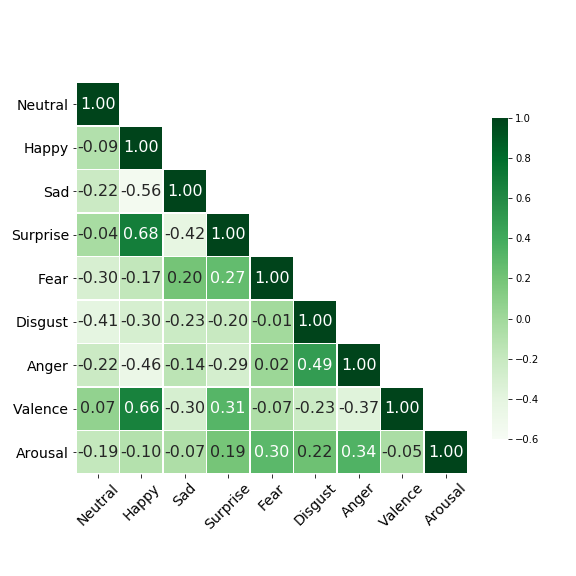}
    \caption{Visualization of the cosine similarity between the learned classifiers and regressors by a GCN that enables the propagation of information between the expression classifiers.}
    \label{fig:similarity_intra}
\end{figure}

To explore the emotional dependencies that the network captures, we visualize the cosine similarity between the learned vectors in Fig. \ref{fig:similarity_intra}. We can see that the similarity between the basic emotions that are adjacent in the graph increases while the similarity between the classifiers and the regressors remains the same as in the Emotion-GCN model. Therefore, the network successfully captured the dependencies that are present between the basic emotions. As stated in the paper, the classification accuracy is slightly lower because we deal with a single-label recognition task and the possible benefits of this approach are more suitable in a multi-label recognition task.

{\small
\bibliographystyle{ieee}
\bibliography{supplementary}
}